%% file: iclr2025WSL_TIPO_arXiv.tex
\documentclass{article} 
\usepackage{iclr2025_conference,times}

\input{math_commands.tex}

\usepackage{hyperref}
\usepackage{url}
\usepackage[capitalize,noabbrev]{cleveref}

\usepackage{graphicx}
\usepackage{subfigure}
\usepackage{color}

\iclrfinalcopy

\title{Can We Optimize Deep RL Policy Weights as Trajectory Modeling?}

\author{Hongyao Tang \thanks{ This work was mainly done by the author in 2022 at Tianjin University.} \\
Mila - Qu\'{e}bec AI Institute\\
Universit\'{e} de Montr\'{e}al\\
\texttt{tang.hongyao}@mila.quebec}

\begin{document}

\maketitle

\begin{abstract}
Learning the optimal policy from a random network initialization is the theme of deep Reinforcement Learning (RL).
As the scale of DRL training increases, treating DRL policy network weights as a new data modality and exploring the potential becomes appealing and possible.
In this work, we focus on the policy learning path in deep RL, represented by \textit{the trajectory of network weights of historical policies}, which reflects the evolvement of the policy learning process.
Taking the idea of trajectory modeling with Transformer, we propose Transformer as Implicit Policy Learner (TIPL), which processes policy network weights in an autoregressive manner.
We collect the policy learning path data by running independent RL training trials, with which we then train our TIPL model.
In the experiments, we demonstrate that TIPL is able to fit the implicit dynamics of policy learning and perform the optimization of policy network by inference.
\end{abstract}

\section{Introduction}
\label{sec:intro}

A lot of achievements of deep learning have been witnessed in the past decade~\citep{MnihKSRVBGRFOPB15DQN,Silver2018AGR,Jumper2021AlphaFold,chatgpt,ONeillRMGPLPGMJ24XEmbodiment}.
With the aim of realizing and surpassing natural intelligence, deep neural network has been the first choice of function approximator to achieve large-scale approximation and generalization.
Despite the success of deep neural network, the understanding of network weights is far from well-known, let alone treating network weights as a data modality and leveraging it to solve problems from a different perspective.

To understand and leverage the weight space of neural network, a consistent effort has been made by the deep learning community in recent years.
To explore the correlation between network weights and network performance, many works study from different angles.
\citet{EilertsenJRUY20Classifying} predicts the hyperparameters of neural network training with network weights.
The correlation between network weights and network performance are studied in terms of accuracy~\citep{Unterthiner2020Predicting}, learning trends~\citep{Martin2020Predicting}, generalization gap~\citep{JiangKMB19PredictingGeneralizationGap,Yak2019Towards}.
Another stream of works focus on learning
implicit neural representations (INRs)~\citep{SchurholtKB21SelfSupervised,NavonSAFCM23Equivariant,LuigiCSRSS23DLINR}.
Comparatively, weight space of neural network is less studied in the context of deep Reinforcement Learning (RL)~\citep{Sutton1988ReinforcementLA}.
As a very first attempt, \citet{Tang2022PeVFA} propose to use the weights of policy network as an additional input of value network to extend the ability.
\citet{Sokar2302Dormant,DohareHLRMS24LossofPlasticity} studies the phenomenon of plasticity loss from the lens of weight dormancy.
Besides, the learning dynamics of an RL agent in the weight space is also studied in~\citep{SchneiderSGCHSB24Identifying,TangB24Ladder}.

In this paper, we study the learning process of a deep RL agent from the lens of weight space.
To be specific, we treat the iterative optimization of agent's policy network (i.e., from randomly initialized weights to the final weights after training) as a trajectory in the weight space, called \textit{policy weight trajectory}.
We then view policy weight trajectory as a new type of data, which reflects how a deep RL agent learns to solve a task with a certain RL algorithm.
Naturally inspired by the appealing trajectory modeling ability of Transformer~\citep{VaswaniSPUJGKP17Transformer}, we aim to study the question: whether Transformer can model policy weight trajectory and take the place of conventional iterative gradient optimization by direct inference.
To this end, we propose \textbf{T}ransformer as \textbf{I}mplicit \textbf{P}olicy
\textbf{L}earner (TIPL), a Transformer-based model for policy weight trajectory, that takes as input the weights of policy network in this history and outputs the weights for the next policy in an autoregressive manner.
In our experiments, we collect policy weight trajectories by running Proximal Policy Optimization (PPO)~\citep{SchulmanWDRK17PPO} in 
two OpenAI Gym MuJoCo continuous control tasks, i.e., InvertedPendulum-v4 and HalfCheetah-v4~\citep{Brockman2016Gym}.
By training TIPL with the collected data, we demonstrate that TIPL is able to fit the implicit dynamics of policy learning and perform the optimization of policy network by inference.

The most related work to our paper is \citep{ZhouYJBXSKF23NeuralFunc}, which proposes to use Transformer to process the network weights for a better INR.
The difference between our work and it is two-fold: first, we process a sequence of network weights rather than the weights of a single network; second, we aim to use Transformer to model the optimization process of the weights in the context of deep RL.
Another related work is Algorithm Distillation~\citep{LaskinWOPSSSHFB23Incontext} where a Transformer is used to fit the cross-episode state-action trajectories of a deep RL agent, consequently showing effective in-context trajectory optimization.
This is different from our work since conventional state-action trajectories are modeled in their work while we focus on the weight trajectory of policy network.

\section{Background}
\label{sec:background}

Reinforcement Learning (RL)~\citep{Sutton1988ReinforcementLA} is a machine learning paradigm of solving the optimal policy for a sequential decision-making problem.
In the framework of RL, the sequential decision-making problem is modeled by a Markov Decision Process (MDP) defined by a tuple $\left< \mathcal{S}, \mathcal{A}, \mathcal{P}, \mathcal{R},  \gamma, \rho_0, T \right>$, with the state set $\mathcal{S}$, the action set $\mathcal{A}$, the transition function $\mathcal{P}: \mathcal{S} \times \mathcal{A} \rightarrow P(\mathcal{S})$, the reward function $\mathcal{R}: \mathcal{S} \times \mathcal{A} \rightarrow \mathbb{R}$, the discounted factor $\gamma \in [0,1)$, the initial state distribution $\rho_0$ and the horizon $T$.
An agent interacts with the MDP by performing actions with its policy $a_t \sim \pi(s_t)$ that defines the mapping from states to action distributions.

The objective of an RL agent is to optimize its policy $\pi$ to maximize the expected discounted cumulative reward
$J(\pi) = \mathbb{E}_{\pi} [\sum_{t=0}^{T}\gamma^{t} r_t ]$,
where $s_{0} \sim \rho_{0}\left(s_{0}\right)$, $s_{t+1} \sim \mathcal{P}\left(s_{t+1} \mid s_{t}, a_{t}\right)$ and $r_t = \mathcal{R}\left(s_{t},a_{t}\right)$.
In deep RL, the agent's policy function is usually approximated with deep neural networks,
conventionally denoted
$\pi_{\phi}$ with network weights $\phi$ (including weights, biases, and etc).
The parameterized policy $\pi_{\phi}$ can be updated by taking the gradient of the objective, 
i.e., $\phi^{\prime} \leftarrow \phi + \alpha \nabla_{\phi} J(\pi_{\phi})$ with a step size $\alpha$~\citep{Silver2014DPG,MnihBMGLHSK16A3C,SchulmanWDRK17PPO,HaarnojaZAL18SAC}.

\section{Method}
\label{sec:method}

In this section, we first introduce the concept of policy weight trajectory (Section~\ref{subsec:policy_path}) and then our Transformer-based model TIPL (Section~\ref{subsec:tipo}).

\subsection{Policy Learning Path as Network Weight Trajectory}
\label{subsec:policy_path}

Starting from a randomly initialized policy network weights $\phi_0$, a canonical learning process of a deep RL policy network is conducted in an iterative manner as $\pi_{\phi_0} \rightarrow \pi_{\phi_1} \rightarrow \dots \rightarrow \pi_{\phi_t} \rightarrow \dots $ within training budget.
In this paper, we call the corresponding sequence of network weights $\mathcal{W} = \{\phi_t\}_{t=0}^{k}$ as \textit{policy weight trajectory}, where $k$ is the max iteration number within budget.

Intuitively, policy weight trajectory reflects the dynamics of the policy optimization process and it is determined by multiple factors including the RL algorithm, the optimizer, the data sampling strategy, random noise, etc.
Formally, the dynamics of policy learning process, or equivalently the distribution of policy weight trajectory can be formulated as
${\phi_t} = \mathbb{P}_{\mathcal{A}}(\pi_{\phi_{t-1}}, D_t, C_t, \epsilon_t)$,
with:
\vspace{-0.2cm}
\begin{itemize}
    \item The RL algorithm $\mathcal{A}$: it determines the calculation of policy gradient.
    \item The data batch $D_t$: it is the data used to compute the gradient sampled by a certain strategy.
    \item The learning context $C_t$: it denotes the context like optimizer status, learning rate schedule and more in a general view.
    \item The random noise $\epsilon_t$: it represents the stochasticity in the sampling and optimization.
\end{itemize}
\vspace{-0.2cm}
In this way, we can view the canonical learning process of deep RL policy network as a process of trajectory generation.
Since Transformer has almost been the first choice in many kinds of complex trajectory modeling problems~\citep{Kirsch2022Generalpurpose}, it is potential to use Transformer to approximate $\mathbb{P}_{\mathcal{A}}$ in our context.
Therefore, it drives us to propose TIPL in the following.

\subsection{Transformer as An Implicit Policy Learner}
\label{subsec:tipo}

To model generation process of policy weight trajectory ${\phi_t} = \mathbb{P}_{\mathcal{A}}(\pi_{\phi_{t-1}}, D_t, C_t, \epsilon_t)$, we adopt GPT~\citep{Radford2018GPT} architecture which has shown its effectiveness in RL literature~\citep{ChenLRLGLASM21DecisionTrans}.
To make the implementation practical and simple, we do not input the factors $D_t, C_t, \epsilon_t$, since it adds a non-trivial burden to maintain these data and inputting them also injects stochasticity to the learning of the Transformer model.
Moreover, we believe that the Transformer model can implicitly infer the information of the missing factors based on the policy weight history $\{\phi_{i}\}_{i=0}^{t-1}$ to some extent.

\begin{figure}[h]
\vspace{-0.3cm}
\centering
\includegraphics[width=0.65\linewidth]{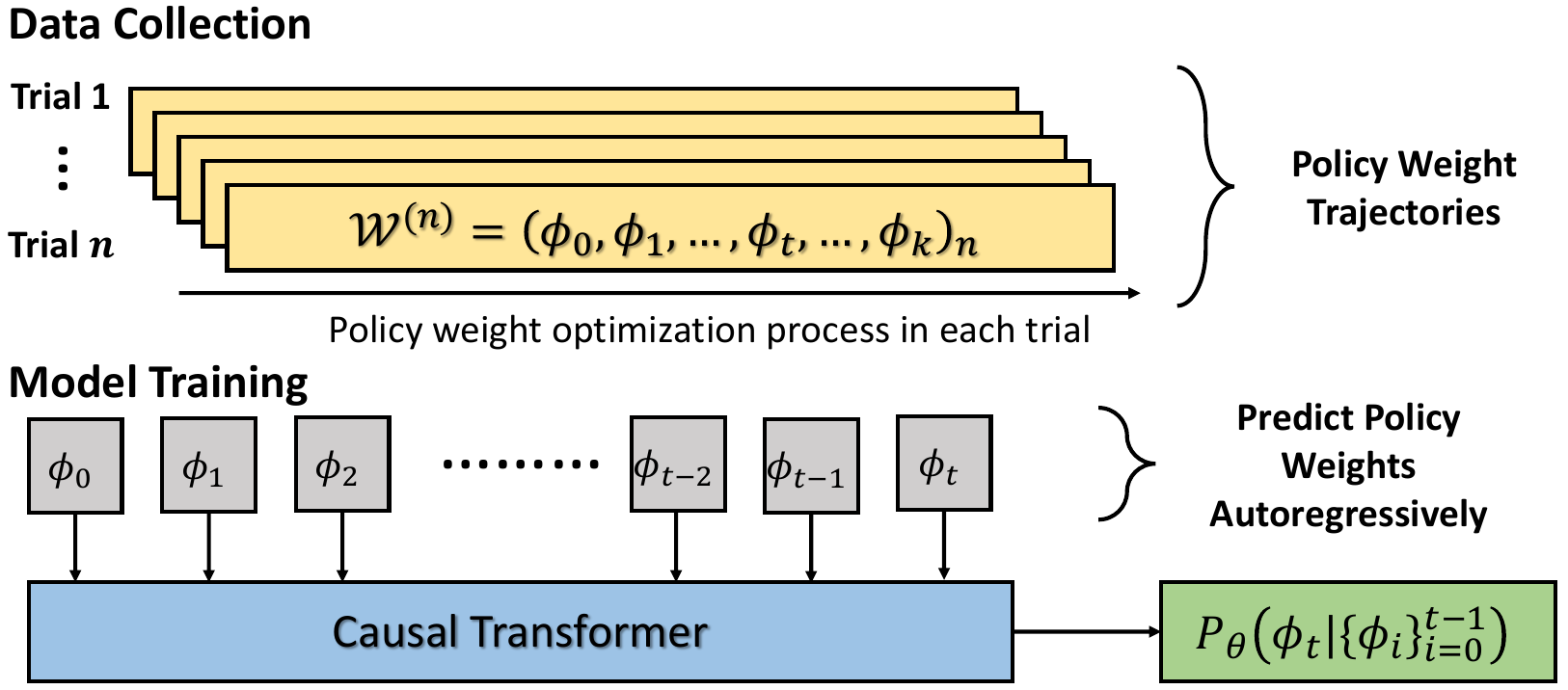}
\vspace{-0.2cm}
\caption{An illustration of Transformer as An Implicit Policy Learner (TIPL).}
\label{figure:tipl_illustration}
\vspace{-0.2cm}
\end{figure}

The illustration of our model, Transformer as An Implicit Policy Learner (TIPL), is shown in Figure~\ref{figure:tipl_illustration}.
TIPL is trained based on the policy weight trajectories $\{\mathcal{W}^{(j)}\}_{j=1}^{n}$ collected from a number of trials of deep RL training.
Each weight $\phi$ in the trajectory $\mathcal{W}$ is viewed as a token.
TIPL takes as input the policy weights $\{\phi_{i}\}_{i=0}^{t-1}$ and outputs the next policy weights $\tilde{\phi}_t \sim P_{\theta}(\cdot | \{\phi_{i}\}_{i=0}^{t-1})$.
TIPL uses a causal Transformer model with a causal mask that only makes the information till timestep $t$ (i.e., $< t$) visible when generating the output at timestep $t$.
Thus, TIPL, parameterized by $\theta$ is trained in autoregressive manner:
\begin{equation}
    \text{minimize}_{\theta} \ \mathbb{E}_{\{{\mathcal{W}^{(j)}\}}_{j=1}^{n}, \tilde{\phi}_t \sim P_{\theta}(\cdot | \{\phi_{i}\}_{i=0}^{t-1})} \ \text{MSE}(\phi_t, \tilde{\phi}_t) .
\end{equation}
Additionally, since a normal-scale RL policy network (e.g., two-layer MLP with 256 neurons per layer) has over 1M weights, it is difficult to directly input it as a token. Instead of learning an implicit network representation, we resort to a simple method, i.e., Temporal SVD~\citep{TangB24Ladder}, to reduce the dimensionality. 
Concretely, by performing SVD decomposition $[\phi_0, \phi_1, \dots, \phi_n]^{\top}=U \Sigma V^{\top}$, the first $d$ columns of the left singular matrix $U$ (denoted by $U_{[:,1:d]}$) is used for a low-dimensional representation $u_i$ of the weight vectors $\phi_i$. This serves as a training-free transformation for each policy weight. In turn, for a $d$-dim representation $\tilde{u}_i$ predicted by TIPL, a full policy weight vector can be reconstructed by $\tilde{\phi}_i = \tilde{u}_i \Sigma_{[1:d,1:d]} V^{\top}_{[:,1:d]}$.

\section{Experiment}
\label{sec:exp}

In this section, we empirically evaluate the performance of TIPL.

\vspace{-0.3cm}
\paragraph{Setups}

In our experiments, we collect policy weight trajectories by running Proximal Policy Optimization (PPO)~\citep{SchulmanWDRK17PPO} in two OpenAI Gym MuJoCo continuous control tasks, i.e., InvertedPendulum-v4 and HalfCheetah-v4~\citep{Brockman2016Gym}.
We adopt the PPO implementation in CleanRL\footnote{\url{https://github.com/vwxyzjn/cleanrl}} with no modification to hyperparameters.
We run 50 independent trials with different random seeds and the total interaction steps are 800k for InvertedPendulum-v4 and 2M for HalfCheetah-v4.
The policy network is a two-layer MLP with 8 or 256 neurons for each layer followed by a conventional Gaussian distribution output, which is sufficient to learn effective policies in the two tasks respectively.
For each trial, we collect the weights of 2k or 5k policies.

We implement TIPL upon the official code of Decision Transformer (DT)~\citep{ChenLRLGLASM21DecisionTrans}\footnote{\url{https://github.com/kzl/decision-transformer}} with minor modifications. The major difference is that we take \textit{policy weights} as input tokens rather than the \textit{(state, action, return-to-go)} inputs in DT. We train TIPL with collected data for 50 iterations with 10k batches of 64 trajectory segments.
We refer readers to the public code repository for specific hyperparameter choices.

\vspace{-0.3cm}
\paragraph{Performance Evaluation}

\begin{figure}
\begin{center}
\vspace{-0.3cm}
\subfigure[Prediction Errors in InvertedPendulum-v4]{
\includegraphics[width=0.3\textwidth]{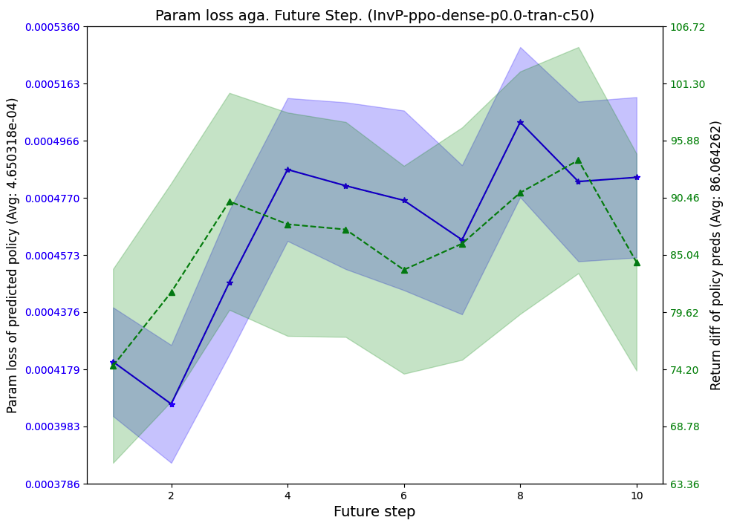}
\includegraphics[width=0.67\textwidth]{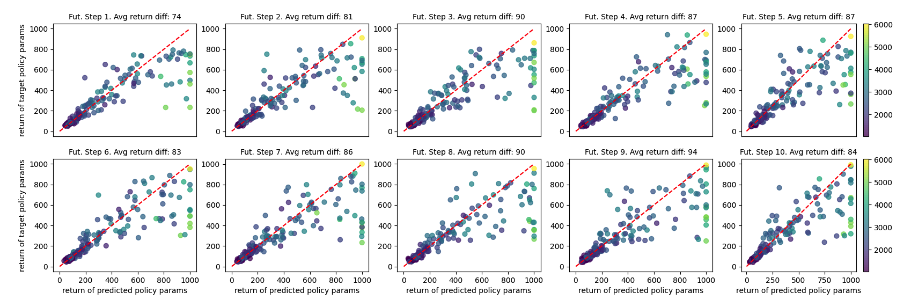}
\vspace{-0.2cm}
}
\subfigure[Prediction Errors in HalfCheetah-v4]{
\includegraphics[width=0.3\textwidth]{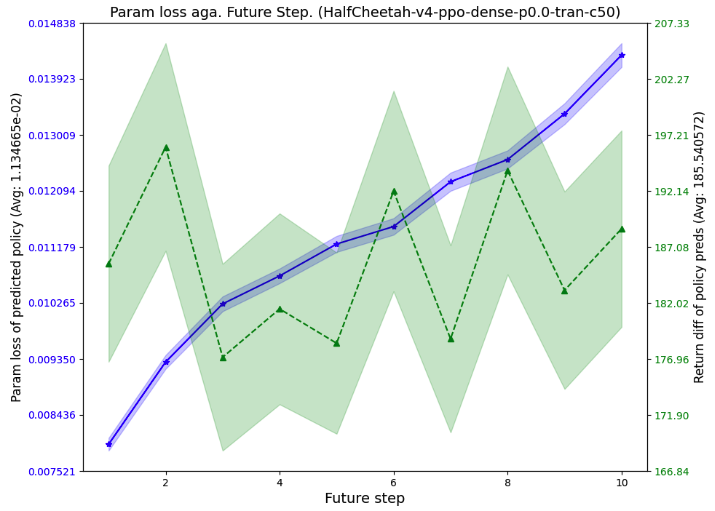}
\includegraphics[width=0.67\textwidth]{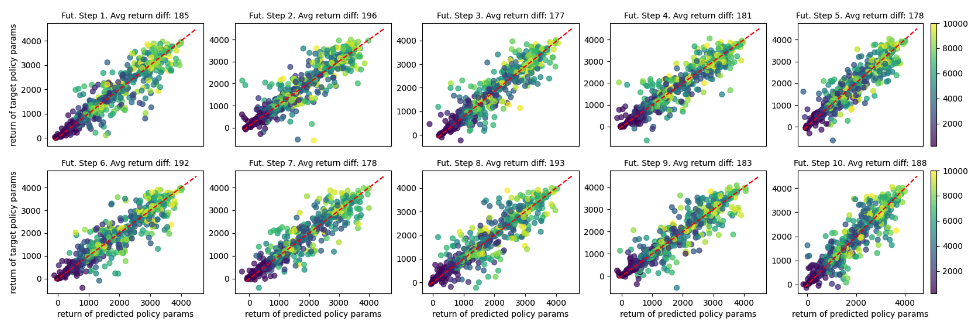}
}
\end{center}
\vspace{-0.4cm}
\caption{Performance evaluation of TIPL. In each panel, the weight prediction error (\textcolor{blue}{blue}) v.s., the return error of predicted weight (\textcolor{green}{green}) is on the left; the scatter plot for return of true policy weight (i.e., $J(\pi_{\phi_t})$) v.s., return of predicted policy weight (i.e., $J(\pi_{\tilde{\phi}_t})$) is on the right.}
\vspace{-0.5cm}
\label{figure:tipl_evaluation}
\end{figure}

To evaluate the performance of TIPL, we use \textit{weight prediction error} (WPE), i.e., $\text{MSE}(\phi_t, \tilde{\phi}_t)$, and \textit{return error of predicted weight} (REPW), i.e., $|J(\pi_{\phi_t}) - J(\pi_{\tilde{\phi}_t})|$.
Note that REPW evaluates whether the predicted weight is indeed an effective policy.
The results are shown in Figure~\ref{figure:tipl_evaluation}.
In the left column, we plot WPE (\textcolor{blue}{blue}) and REPW (\textcolor{green}{green}) against the forward prediction steps, i.e., $\tilde{\phi}_{t}, \tilde{\phi}_{t+1}, \tilde{\phi}_{t+2},
\dots$.
We can observe that TIPL shows higher WPE as the forward prediction step increases; while REPW basically stays at the same level.
This means that the compounding WPE in predicted weights does not lead to deteriorating policies, as minor differences in weight space may not influence policy performance.
In the right column, we compare the return of true policy weight $J(\pi_{\phi_t})$ and the return of predicted policy weight $J(\pi_{\tilde{\phi}_t})$ with scatter points across ten forward prediction steps.
We can see that the scatter points lie around the diagonal line.
This indicates that TIPL is able to capture the overall improvement trend of policy weight trajectory.
Note that REPW should not be fully attributed to TIPL, as it is also influenced by the stochasticity in the return evaluation of policy network and gradient optimization.
In summary, we demonstrate that TIPL is able to model the evolvement of deep RL policy in weight space and generate the next policy weight based on historical weights in our experiments.

\section{Conclusion and Discussion}
\label{sec:conclusion}

In this paper, we explore the idea of using Transformer to autoregressively approximate the policy weight trajectory generated by typical RL training processes.
Our experiments demonstrate the effectiveness of TIPL in predicting the weight of improved policies.

\vspace{-0.3cm}
\paragraph{Policy Data Limitation} 
A large amount of data is one of the prerequisites for eliciting the power of a large Transformer model. Policy weight is a type of highly dense data that encapsulates the decision-making behaviors of an RL agent. We expect to build a larger dataset for policy weight trajectory so as to further our study by better leveraging the power of Transformer.

\vspace{-0.3cm}
\paragraph{Better Architecture and Weight Representation} We adopt GPT architecture and use Temporal SVD for a simple dimension reduction for policy weight. Considering more sophisticated implicit neural representation and advanced Transformer architecture especially tailored for network weight is a straightforward future direction.


\vspace{-0.3cm}
\paragraph{Usage of TIPL} With the scaling of policy weight trajectory data and TIPL model, we expect that TIPL can be used as a general optimizer for deep RL policy across different learning problems and reduce the reliance on iterative gradient optimization.



\bibliography{iclr2025_conference}
\bibliographystyle{iclr2025_conference}


\end{document}

%% file: math_commands.tex
\usepackage{amsmath,amsfonts,bm}

\def\eqref#1{equation~\ref{#1}}

\def\1{\bm{1}}

\DeclareMathAlphabet{\mathsfit}{\encodingdefault}{\sfdefault}{m}{sl}
\SetMathAlphabet{\mathsfit}{bold}{\encodingdefault}{\sfdefault}{bx}{n}













%% file: iclr2025WSL_TIPO_arXiv.bbl
\begin{thebibliography}{30}
\providecommand{\natexlab}[1]{#1}
\providecommand{\url}[1]{\texttt{#1}}
\expandafter\ifx\csname urlstyle\endcsname\relax
  \providecommand{\doi}[1]{doi: #1}\else
  \providecommand{\doi}{doi: \begingroup \urlstyle{rm}\Url}\fi

\bibitem[Brockman et~al.(2016)Brockman, Cheung, Pettersson, Schneider, Schulman, Tang, and Zaremba]{Brockman2016Gym}
G.~Brockman, V.~Cheung, L.~Pettersson, J.~Schneider, J.~Schulman, J.~Tang, and W.~Zaremba.
\newblock Openai gym.
\newblock \emph{arXiv preprint}, arXiv:1606.01540, 2016.

\bibitem[Chen et~al.(2021)Chen, Lu, Rajeswaran, Lee, Grover, Laskin, Abbeel, Srinivas, and Mordatch]{ChenLRLGLASM21DecisionTrans}
Lili Chen, Kevin Lu, Aravind Rajeswaran, Kimin Lee, Aditya Grover, Michael Laskin, Pieter Abbeel, Aravind Srinivas, and Igor Mordatch.
\newblock Decision transformer: Reinforcement learning via sequence modeling.
\newblock In \emph{NeurIPS}, pp.\  15084--15097, 2021.

\bibitem[Dohare et~al.(2024)Dohare, Hernandez{-}Garcia, Lan, Rahman, Mahmood, and Sutton]{DohareHLRMS24LossofPlasticity}
Shibhansh Dohare, J.~Fernando Hernandez{-}Garcia, Qingfeng Lan, Parash Rahman, A.~Rupam Mahmood, and Richard~S. Sutton.
\newblock Loss of plasticity in deep continual learning.
\newblock \emph{Nature}, 632\penalty0 (8026):\penalty0 768--774, 2024.

\bibitem[Eilertsen et~al.(2020)Eilertsen, J{\"{o}}nsson, Ropinski, Unger, and Ynnerman]{EilertsenJRUY20Classifying}
Gabriel Eilertsen, Daniel J{\"{o}}nsson, Timo Ropinski, Jonas Unger, and Anders Ynnerman.
\newblock Classifying the classifier: Dissecting the weight space of neural networks.
\newblock In \emph{ECAI}, volume 325, pp.\  1119--1126, 2020.

\bibitem[Haarnoja et~al.(2018)Haarnoja, Zhou, Abbeel, and Levine]{HaarnojaZAL18SAC}
T.~Haarnoja, A.~Zhou, P.~Abbeel, and S.~Levine.
\newblock Soft actor-critic: Off-policy maximum entropy deep reinforcement learning with a stochastic actor.
\newblock In \emph{ICML}, 2018.

\bibitem[Jiang et~al.(2019)Jiang, Krishnan, Mobahi, and Bengio]{JiangKMB19PredictingGeneralizationGap}
Yiding Jiang, Dilip Krishnan, Hossein Mobahi, and Samy Bengio.
\newblock Predicting the generalization gap in deep networks with margin distributions.
\newblock In \emph{ICLR}, 2019.

\bibitem[Jumper et~al.(2021)Jumper, Evans, Pritzel, Green, Figurnov, Ronneberger, Tunyasuvunakool, Bates, Ž{\'i}dek, Potapenko, Bridgland, Meyer, Kohl, Ballard, Cowie, Romera-Paredes, Nikolov, Jain, Adler, Back, Petersen, Reiman, Clancy, Zielinski, Steinegger, Pacholska, Berghammer, Bodenstein, Silver, Vinyals, Senior, Kavukcuoglu, Kohli, and Hassabis]{Jumper2021AlphaFold}
John~M. Jumper, Richard Evans, Alexander Pritzel, Tim Green, Michael Figurnov, Olaf Ronneberger, Kathryn Tunyasuvunakool, Russ Bates, Augustin Ž{\'i}dek, Anna Potapenko, Alex Bridgland, Clemens Meyer, Simon A~A Kohl, Andy Ballard, Andrew Cowie, Bernardino Romera-Paredes, Stanislav Nikolov, Rishub Jain, Jonas Adler, Trevor Back, Stig Petersen, David Reiman, Ellen Clancy, Michal Zielinski, Martin Steinegger, Michalina Pacholska, Tamas Berghammer, Sebastian Bodenstein, David Silver, Oriol Vinyals, Andrew~W. Senior, Koray Kavukcuoglu, Pushmeet Kohli, and Demis Hassabis.
\newblock Highly accurate protein structure prediction with alphafold.
\newblock \emph{Nature}, 596:\penalty0 583 -- 589, 2021.

\bibitem[Kirsch et~al.(2022)Kirsch, Harrison, Sohl{-}Dickstein, and Metz]{Kirsch2022Generalpurpose}
Louis Kirsch, James Harrison, Jascha Sohl{-}Dickstein, and Luke Metz.
\newblock General-purpose in-context learning by meta-learning transformers.
\newblock \emph{arXiv preprint, arXiv:2212.04458}, 2022.

\bibitem[Laskin et~al.(2023)Laskin, Wang, Oh, Parisotto, Spencer, Steigerwald, Strouse, Hansen, Filos, Brooks, Gazeau, Sahni, Singh, and Mnih]{LaskinWOPSSSHFB23Incontext}
Michael Laskin, Luyu Wang, Junhyuk Oh, Emilio Parisotto, Stephen Spencer, Richie Steigerwald, DJ~Strouse, Steven~Stenberg Hansen, Angelos Filos, Ethan Brooks, Maxime Gazeau, Himanshu Sahni, Satinder Singh, and Volodymyr Mnih.
\newblock In-context reinforcement learning with algorithm distillation.
\newblock In \emph{ICLR}, 2023.

\bibitem[Luigi et~al.(2023)Luigi, Cardace, Spezialetti, Ramirez, Salti, and Stefano]{LuigiCSRSS23DLINR}
Luca~De Luigi, Adriano Cardace, Riccardo Spezialetti, Pierluigi~Zama Ramirez, Samuele Salti, and Luigi~Di Stefano.
\newblock Deep learning on implicit neural representations of shapes.
\newblock In \emph{ICLR}, 2023.

\bibitem[Martin et~al.(2020)Martin, Peng, and Mahoney]{Martin2020Predicting}
Charles~H. Martin, Tongsu Peng, and Michael~W. Mahoney.
\newblock Predicting trends in the quality of state-of-the-art neural networks without access to training or testing data.
\newblock \emph{arXiv preprint, arXiv:2002.06716}, 2020.

\bibitem[Mnih et~al.(2015)Mnih, Kavukcuoglu, Silver, Rusu, Veness, Bellemare, Graves, Riedmiller, Fidjeland, Ostrovski, Petersen, Beattie, Sadik, Antonoglou, King, Kumaran, Wierstra, Legg, and Hassabis]{MnihKSRVBGRFOPB15DQN}
Volodymyr Mnih, Koray Kavukcuoglu, David Silver, Andrei~A. Rusu, Joel Veness, Marc~G. Bellemare, Alex Graves, Martin~A. Riedmiller, Andreas Fidjeland, Georg Ostrovski, Stig Petersen, Charles Beattie, Amir Sadik, Ioannis Antonoglou, Helen King, Dharshan Kumaran, Daan Wierstra, Shane Legg, and Demis Hassabis.
\newblock Human-level control through deep reinforcement learning.
\newblock \emph{Nature}, 518\penalty0 (7540):\penalty0 529--533, 2015.

\bibitem[Mnih et~al.(2016)Mnih, Badia, Mirza, Graves, Lillicrap, Harley, Silver, and Kavukcuoglu]{MnihBMGLHSK16A3C}
Volodymyr Mnih, Adri{\`{a}}~Puigdom{\`{e}}nech Badia, Mehdi Mirza, Alex Graves, Timothy~P. Lillicrap, Tim Harley, David Silver, and Koray Kavukcuoglu.
\newblock Asynchronous methods for deep reinforcement learning.
\newblock In \emph{ICML}, volume~48, pp.\  1928--1937, 2016.

\bibitem[Navon et~al.(2023)Navon, Shamsian, Achituve, Fetaya, Chechik, and Maron]{NavonSAFCM23Equivariant}
Aviv Navon, Aviv Shamsian, Idan Achituve, Ethan Fetaya, Gal Chechik, and Haggai Maron.
\newblock Equivariant architectures for learning in deep weight spaces.
\newblock In \emph{ICML}, volume 202 of \emph{Proceedings of Machine Learning Research}, pp.\  25790--25816, 2023.

\bibitem[O'Neill et~al.(2024)O'Neill, Rehman, Maddukuri, Gupta, Padalkar, Lee, Pooley, Gupta, Mandlekar, Jain, Tung, Bewley, Herzog, Irpan, Khazatsky, Rai, Gupta, Wang, Singh, Garg, Kembhavi, Xie, Brohan, Raffin, Sharma, Yavary, Jain, Balakrishna, Wahid, Burgess{-}Limerick, Kim, Sch{\"{o}}lkopf, Wulfe, Ichter, Lu, Xu, Le, Finn, Wang, Xu, Chi, Huang, Chan, Agia, Pan, Fu, Devin, Xu, Morton, Driess, Chen, Pathak, Shah, B{\"{u}}chler, Jayaraman, Kalashnikov, Sadigh, Johns, Foster, Liu, Ceola, Xia, Zhao, Stulp, Zhou, Sukhatme, Salhotra, Yan, Feng, Schiavi, Berseth, Kahn, Wang, Su, Fang, Shi, Bao, Amor, Christensen, Furuta, Walke, Fang, Ha, Mordatch, Radosavovic, Leal, Liang, Abou{-}Chakra, Kim, Drake, Peters, Schneider, Hsu, Bohg, Bingham, Wu, Gao, Hu, Wu, Wu, Sun, Luo, Gu, Tan, Oh, Wu, Lu, Yang, Malik, Silv{\'{e}}rio, Hejna, Booher, Tompson, Yang, Salvador, Lim, Han, Wang, Rao, Pertsch, Hausman, Go, Gopalakrishnan, Goldberg, Byrne, Oslund, Kawaharazuka, Black, Lin, Zhang, Ehsani, Lekkala, Ellis, Rana, Srinivasan,
  Fang, Singh, Zeng, Hatch, Hsu, Itti, Chen, Pinto, Fei{-}Fei, Tan, Fan, Ott, Lee, Weihs, Chen, Lepert, Memmel, Tomizuka, Itkina, Castro, Spero, Du, Ahn, Yip, Zhang, Ding, Heo, Srirama, Sharma, Kim, Kanazawa, Hansen, Heess, Joshi, S{\"{u}}nderhauf, Liu, Palo, Shafiullah, Mees, Kroemer, Bastani, Sanketi, Miller, Yin, Wohlhart, Xu, Fagan, Mitrano, Sermanet, Abbeel, Sundaresan, Chen, Vuong, Rafailov, Tian, Doshi, Mart{\'{\i}}n{-}Mart{\'{\i}}n, Baijal, Scalise, Hendrix, Lin, Qian, Zhang, Mendonca, Shah, Hoque, Julian, Bustamante, Kirmani, Levine, Lin, Moore, Bahl, Dass, Sonawani, Song, Xu, Haldar, Karamcheti, Adebola, Guist, Nasiriany, Schaal, Welker, Tian, Ramamoorthy, Dasari, Belkhale, Park, Nair, Mirchandani, Osa, Gupta, Harada, Matsushima, Xiao, Kollar, Yu, Ding, Davchev, Zhao, Armstrong, Darrell, Chung, Jain, Vanhoucke, Zhan, Zhou, Burgard, Chen, Wang, Zhu, Geng, Liu, Xu, Li, Lu, Ma, Kim, Chebotar, Zhou, Zhu, Wu, Xu, Wang, Bisk, Cho, Lee, Cui, Cao, Wu, Tang, Zhu, Zhang, Jiang, Li, Li, Iwasawa, Matsuo, Ma,
  Xu, Cui, Zhang, and Lin]{ONeillRMGPLPGMJ24XEmbodiment}
Abby O'Neill, Abdul Rehman, Abhiram Maddukuri, Abhishek Gupta, Abhishek Padalkar, Abraham Lee, Acorn Pooley, Agrim Gupta, Ajay Mandlekar, Ajinkya Jain, Albert Tung, Alex Bewley, Alexander Herzog, Alex Irpan, Alexander Khazatsky, Anant Rai, Anchit Gupta, Andrew Wang, Anikait Singh, Animesh Garg, Aniruddha Kembhavi, Annie Xie, Anthony Brohan, Antonin Raffin, Archit Sharma, Arefeh Yavary, Arhan Jain, Ashwin Balakrishna, Ayzaan Wahid, Ben Burgess{-}Limerick, Beomjoon Kim, Bernhard Sch{\"{o}}lkopf, Blake Wulfe, Brian Ichter, Cewu Lu, Charles Xu, Charlotte Le, Chelsea Finn, Chen Wang, Chenfeng Xu, Cheng Chi, Chenguang Huang, Christine Chan, Christopher Agia, Chuer Pan, Chuyuan Fu, Coline Devin, Danfei Xu, Daniel Morton, Danny Driess, Daphne Chen, Deepak Pathak, Dhruv Shah, Dieter B{\"{u}}chler, Dinesh Jayaraman, Dmitry Kalashnikov, Dorsa Sadigh, Edward Johns, Ethan~Paul Foster, Fangchen Liu, Federico Ceola, Fei Xia, Feiyu Zhao, Freek Stulp, Gaoyue Zhou, Gaurav~S. Sukhatme, Gautam Salhotra, Ge~Yan, Gilbert Feng,
  Giulio Schiavi, Glen Berseth, Gregory Kahn, Guanzhi Wang, Hao Su, Haoshu Fang, Haochen Shi, Henghui Bao, Heni~Ben Amor, Henrik~I. Christensen, Hiroki Furuta, Homer Walke, Hongjie Fang, Huy Ha, Igor Mordatch, Ilija Radosavovic, Isabel Leal, Jacky Liang, Jad Abou{-}Chakra, Jaehyung Kim, Jaimyn Drake, Jan Peters, Jan Schneider, Jasmine Hsu, Jeannette Bohg, Jeffrey Bingham, Jeffrey Wu, Jensen Gao, Jiaheng Hu, Jiajun Wu, Jialin Wu, Jiankai Sun, Jianlan Luo, Jiayuan Gu, Jie Tan, Jihoon Oh, Jimmy Wu, Jingpei Lu, Jingyun Yang, Jitendra Malik, Jo{\~{a}}o Silv{\'{e}}rio, Joey Hejna, Jonathan Booher, Jonathan Tompson, Jonathan Yang, Jordi Salvador, Joseph~J. Lim, Junhyek Han, Kaiyuan Wang, Kanishka Rao, Karl Pertsch, Karol Hausman, Keegan Go, Keerthana Gopalakrishnan, Ken Goldberg, Kendra Byrne, Kenneth Oslund, Kento Kawaharazuka, Kevin Black, Kevin Lin, Kevin Zhang, Kiana Ehsani, Kiran Lekkala, Kirsty Ellis, Krishan Rana, Krishnan Srinivasan, Kuan Fang, Kunal~Pratap Singh, Kuo{-}Hao Zeng, Kyle Hatch, Kyle Hsu,
  Laurent Itti, Lawrence~Yunliang Chen, Lerrel Pinto, Li~Fei{-}Fei, Liam Tan, Linxi~Jim Fan, Lionel Ott, Lisa Lee, Luca Weihs, Magnum Chen, Marion Lepert, Marius Memmel, Masayoshi Tomizuka, Masha Itkina, Mateo~Guaman Castro, Max Spero, Maximilian Du, Michael Ahn, Michael~C. Yip, Mingtong Zhang, Mingyu Ding, Minho Heo, Mohan~Kumar Srirama, Mohit Sharma, Moo~Jin Kim, Naoaki Kanazawa, Nicklas Hansen, Nicolas Heess, Nikhil~J. Joshi, Niko S{\"{u}}nderhauf, Ning Liu, Norman~Di Palo, Nur Muhammad~(Mahi) Shafiullah, Oier Mees, Oliver Kroemer, Osbert Bastani, Pannag~R. Sanketi, Patrick~Tree Miller, Patrick Yin, Paul Wohlhart, Peng Xu, Peter~David Fagan, Peter Mitrano, Pierre Sermanet, Pieter Abbeel, Priya Sundaresan, Qiuyu Chen, Quan Vuong, Rafael Rafailov, Ran Tian, Ria Doshi, Roberto Mart{\'{\i}}n{-}Mart{\'{\i}}n, Rohan Baijal, Rosario Scalise, Rose Hendrix, Roy Lin, Runjia Qian, Ruohan Zhang, Russell Mendonca, Rutav Shah, Ryan Hoque, Ryan Julian, Samuel Bustamante, Sean Kirmani, Sergey Levine, Shan Lin, Sherry
  Moore, Shikhar Bahl, Shivin Dass, Shubham~D. Sonawani, Shuran Song, Sichun Xu, Siddhant Haldar, Siddharth Karamcheti, Simeon Adebola, Simon Guist, Soroush Nasiriany, Stefan Schaal, Stefan Welker, Stephen Tian, Subramanian Ramamoorthy, Sudeep Dasari, Suneel Belkhale, Sungjae Park, Suraj Nair, Suvir Mirchandani, Takayuki Osa, Tanmay Gupta, Tatsuya Harada, Tatsuya Matsushima, Ted Xiao, Thomas Kollar, Tianhe Yu, Tianli Ding, Todor Davchev, Tony~Z. Zhao, Travis Armstrong, Trevor Darrell, Trinity Chung, Vidhi Jain, Vincent Vanhoucke, Wei Zhan, Wenxuan Zhou, Wolfram Burgard, Xi~Chen, Xiaolong Wang, Xinghao Zhu, Xinyang Geng, Xiyuan Liu, Liangwei Xu, Xuanlin Li, Yao Lu, Yecheng~Jason Ma, Yejin Kim, Yevgen Chebotar, Yifan Zhou, Yifeng Zhu, Yilin Wu, Ying Xu, Yixuan Wang, Yonatan Bisk, Yoonyoung Cho, Youngwoon Lee, Yuchen Cui, Yue Cao, Yueh{-}Hua Wu, Yujin Tang, Yuke Zhu, Yunchu Zhang, Yunfan Jiang, Yunshuang Li, Yunzhu Li, Yusuke Iwasawa, Yutaka Matsuo, Zehan Ma, Zhuo Xu, Zichen~Jeff Cui, Zichen Zhang, and Zipeng
  Lin.
\newblock Open x-embodiment: Robotic learning datasets and {RT-X} models : Open x-embodiment collaboration.
\newblock In \emph{ICRA}, pp.\  6892--6903, 2024.

\bibitem[OpenAI(2022)]{chatgpt}
OpenAI.
\newblock Chatgpt: Optimizing language models for dialogue, 2022.
\newblock URL \url{https://openai.com/blog/chatgpt/}.

\bibitem[Radford \& Narasimhan(2018)Radford and Narasimhan]{Radford2018GPT}
Alec Radford and Karthik Narasimhan.
\newblock Improving language understanding by generative pre-training.
\newblock 2018.

\bibitem[Schneider et~al.(2024)Schneider, Schumacher, Guist, Chen, Haeufle, Sch{\"{o}}lkopf, and B{\"{u}}chler]{SchneiderSGCHSB24Identifying}
Jan Schneider, Pierre Schumacher, Simon Guist, Le~Chen, Daniel F.~B. Haeufle, Bernhard Sch{\"{o}}lkopf, and Dieter B{\"{u}}chler.
\newblock Identifying policy gradient subspaces.
\newblock In \emph{ICLR}, 2024.

\bibitem[Schulman et~al.(2017)Schulman, Wolski, Dhariwal, Radford, and Klimov]{SchulmanWDRK17PPO}
John Schulman, Filip Wolski, Prafulla Dhariwal, Alec Radford, and Oleg Klimov.
\newblock Proximal policy optimization algorithms.
\newblock \emph{arXiv preprint}, arXiv:1707.06347, 2017.

\bibitem[Sch{\"{u}}rholt et~al.(2021)Sch{\"{u}}rholt, Kostadinov, and Borth]{SchurholtKB21SelfSupervised}
Konstantin Sch{\"{u}}rholt, Dimche Kostadinov, and Damian Borth.
\newblock Self-supervised representation learning on neural network weights for model characteristic prediction.
\newblock In \emph{NeurIPS}, pp.\  16481--16493, 2021.

\bibitem[Silver et~al.(2014)Silver, Lever, Heess, Degris, Wierstra, and Riedmiller]{Silver2014DPG}
D.~Silver, G.~Lever, N.~Heess, T.~Degris, D.~Wierstra, and M.~A. Riedmiller.
\newblock Deterministic policy gradient algorithms.
\newblock In \emph{ICML}, 2014.

\bibitem[Silver et~al.(2018)Silver, Hubert, Schrittwieser, Antonoglou, Lai, Guez, Lanctot, Sifre, Kumaran, Graepel, Lillicrap, Simonyan, and Hassabis]{Silver2018AGR}
David Silver, Thomas Hubert, Julian Schrittwieser, Ioannis Antonoglou, Matthew Lai, Arthur Guez, Marc Lanctot, L.~Sifre, Dharshan Kumaran, Thore Graepel, Timothy~P. Lillicrap, Karen Simonyan, and Demis Hassabis.
\newblock A general reinforcement learning algorithm that masters chess, shogi, and go through self-play.
\newblock \emph{Science}, 362:\penalty0 1140 -- 1144, 2018.

\bibitem[Sokar et~al.(2023)Sokar, Agarwal, Castro, and Evci]{Sokar2302Dormant}
Ghada Sokar, Rishabh Agarwal, Pablo~Samuel Castro, and Utku Evci.
\newblock The dormant neuron phenomenon in deep reinforcement learning.
\newblock In \emph{ICML}, pp.\  32145--32168, 2023.

\bibitem[Sutton \& Barto(1988)Sutton and Barto]{Sutton1988ReinforcementLA}
R.~S. Sutton and A.~G. Barto.
\newblock Reinforcement learning: An introduction.
\newblock \emph{IEEE Transactions on Neural Networks}, 16:\penalty0 285--286, 1988.

\bibitem[Tang et~al.(2022)Tang, Meng, Hao, Chen, Graves, Li, Yu, Mao, Liu, Yang, Tao, and Wang]{Tang2022PeVFA}
Hongyao Tang, Zhaopeng Meng, Jianye Hao, Chen Chen, Daniel Graves, Dong Li, Changmin Yu, Hangyu Mao, Wulong Liu, Yaodong Yang, Wenyuan Tao, and Li~Wang.
\newblock What about inputting policy in value function: Policy representation and policy-extended value function approximator.
\newblock In \emph{AAAI}, pp.\  8441--8449, 2022.

\bibitem[Tang et~al.(2024)Tang, Zhang, Chen, and Hao]{TangB24Ladder}
Hongyao Tang, Min Zhang, Chen Chen, and Jianye Hao.
\newblock The ladder in chaos: Improving policy learning by harnessing the parameter evolving path in a low-dimensional space.
\newblock In \emph{NeurIPS}, 2024.

\bibitem[Unterthiner et~al.(2020)Unterthiner, Keysers, Gelly, Bousquet, and Tolstikhin]{Unterthiner2020Predicting}
Thomas Unterthiner, Daniel Keysers, Sylvain Gelly, Olivier Bousquet, and Ilya~O. Tolstikhin.
\newblock Predicting neural network accuracy from weights.
\newblock \emph{arXiv preprint, arXiv:2002.11448}, 2020.

\bibitem[Vaswani et~al.(2017)Vaswani, Shazeer, Parmar, Uszkoreit, Jones, Gomez, Kaiser, and Polosukhin]{VaswaniSPUJGKP17Transformer}
Ashish Vaswani, Noam Shazeer, Niki Parmar, Jakob Uszkoreit, Llion Jones, Aidan~N. Gomez, Lukasz Kaiser, and Illia Polosukhin.
\newblock Attention is all you need.
\newblock In \emph{NeurIPS}, pp.\  5998--6008, 2017.

\bibitem[Yak et~al.(2019)Yak, Gonzalvo, and Mazzawi]{Yak2019Towards}
Scott Yak, Javier Gonzalvo, and Hanna Mazzawi.
\newblock Towards task and architecture-independent generalization gap predictors.
\newblock \emph{arXiv preprint, arXiv:1906.01550}, 2019.

\bibitem[Zhou et~al.(2023)Zhou, Yang, Jiang, Burns, Xu, Sokota, Kolter, and Finn]{ZhouYJBXSKF23NeuralFunc}
Allan Zhou, Kaien Yang, Yiding Jiang, Kaylee Burns, Winnie Xu, Samuel Sokota, J.~Zico Kolter, and Chelsea Finn.
\newblock Neural functional transformers.
\newblock In \emph{NeurIPS}, 2023.

\end{thebibliography}
